\documentclass[letterpaper, 10 pt, conference]{ieeeconf}  

\IEEEoverridecommandlockouts                              
\usepackage{epsfig}
\makeatletter
\let\NAT@parse\undefined
\makeatother
\usepackage{comment}
\usepackage{epstopdf}
\usepackage{algorithm}
\usepackage{algpseudocode}
\usepackage{flushend}
\usepackage{amsfonts}
\usepackage{cite}
\usepackage{epstopdf}
\usepackage{multicol}
\usepackage{graphicx}

\usepackage{nicefrac}
\usepackage{harpoon}
\usepackage{amssymb} 
\usepackage{amsmath,xparse} 
\usepackage{wrapfig,lipsum}
\usepackage{epstopdf}
\usepackage{epsfig}
\usepackage{array}
\newcolumntype{P}[1]{>{\centering\arraybackslash}p{#1}}
\newcolumntype{M}[1]{>{\centering\arraybackslash}m{#1}}
\usepackage{amsmath}

\usepackage{epsfig}
\makeatletter
\let\NAT@parse\undefined
\makeatother
\usepackage{footnote}
\makesavenoteenv{tabular}
\makesavenoteenv{table}
\usepackage{booktabs}       
\usepackage{comment}
\usepackage{epstopdf}
\usepackage{algorithm}
\usepackage{algpseudocode}
\usepackage{flushend}
\usepackage{amsfonts}
\usepackage{cite}
\usepackage{epstopdf}
\usepackage[font={footnotesize}]{caption}
\usepackage{multicol}
\usepackage{graphicx}

\usepackage{subcaption}  
\usepackage{nicefrac}
\usepackage{amssymb} 
\usepackage{amsmath,xparse} 
\usepackage{wrapfig,lipsum}
\usepackage{epstopdf}
\usepackage{epsfig}

\usepackage{array}

\usepackage{harpoon}

\title{\LARGE \bf
Real-time Data Driven Precision Estimator for RAVEN-II Surgical Robot End Effector Position
}

\author{Haonan Peng, Xingjian Yang, Yun-Hsuan Su, Blake Hannaford
\thanks{Haonan Peng, Xingjian Yang, Yun-Hsuan Su and Blake Hannaford are with the University of Washington Department of Electrical and Computer Engineering, 185 Stevens Way, Paul Allen Center - Room AE100R, Campus Box 352500, Seattle, WA 98195-2500, USA.
{\tt\small \{penghn, yxj1995, yhsu83, blake\}@uw.edu}}
\thanks{978-1-5386-2512-5/18/\textdollar31.00 \copyright2020 IEEE}}

\begin{document}

\maketitle
\thispagestyle{empty}
\pagestyle{empty}

\begin{abstract}
Surgical robots have been introduced to operating rooms over the past few decades due to their high sensitivity, small size, and remote controllability. The cable-driven nature of many surgical robots allows the systems to be dexterous and lightweight, with diameters as low as 5mm. However, due to the slack and stretch of the cables and the backlash of the gears, inevitable uncertainties are brought into the kinematics calculation \cite{haghighipanah2015improving}. Since the reported end effector position of surgical robots like RAVEN-II \cite{hannaford2012raven} is directly calculated using the motor encoder measurements and forward kinematics, it may contain relatively large error up to 10mm, whereas semi-autonomous functions being introduced into abdominal surgeries require position inaccuracy of at most 1mm. To resolve the problem, a cost-effective, real-time and data-driven pipeline for robot end effector position precision estimation is proposed and tested on RAVEN-II. Analysis shows an improved end effector position error of around 1mm RMS traversing through the entire robot workspace without high-resolution motion tracker. The open source code, data sets, videos, and user guide can be found at //github.com/HaonanPeng/RAVEN\_Neural\_Network\_Estimator.
\end{abstract}

\section{Introduction}
\subsection{Background}
Robot-assisted Minimally Invasive Surgery (RAMIS) opens the door to collaborative operations between experienced surgeons and surgical robots with high dexterity and robustness \cite{palep2009robotic}. While surgeons are in charge of decision making and robot manipulation through teleoperation, robots follow the trajectory commands. In abdominal RAMIS, the precision requirement is in millimeter scale. With surgeons manually closing the loop, accuracy of the reported robot end effector pose is not a big concern. In recent years, surgical robot intelligence has emerged in medical robotics research, where repetitive tasks like ablation \cite{hu2015semi} and debridement \cite{kehoe2014autonomous} can be conducted autonomously under supervision of surgeons. Intelligent robot navigation agents are now being developed to incorporate raw teleoperation commands, tremor canceling \cite{riviere2003toward} and motion compensation of the dynamic surgical scenes \cite{lindgren2017towards}\cite{yuen2009robotic}. Moreover, vision-based force estimation \cite{ismr2018} in RAMIS shows promise. In all these applications, precise end effector positioning of the surgical robot is a requirement. 

Many surgical robots are designed with cable transmissions with motors mounted at the base to allow lighter and more compact arms. Cable dynamic properties such as stiffness and internal damping are significant and are known to vary as a function of tension \cite{kosari2013control}. 
Therefore, the kinematics based end effector poses reported by the robot from motor sensors are prone to error.

\subsection{Related Work}
To compensate for the inaccuracy, an intuitive approach is to take additional sensor measurements by applying a motion tracker at the surgical tooltip or a joint encoder on each robot joint depending on whether to resolve the problem in cartesian or joint level. Drawbacks occur in both solutions. In the former case, complications occur during the required high heat sterilization procedure \cite{kosari2013control}. The latter, however, introduces complexity keeping the sensor wires and robot cables compact. Alternatively, real-time video streams from endoscopes are used as an additional cue for end effector pose estimates. But extracting pose information from vision alone can be challenging with the highly dynamic and reflective surgical scenes in real-world operations \cite{lin2016video}. Recently, online estimation systems are proposed to provide a robust and precise end effector position prediction. 

\begin{figure*}[t]
  \centering
  \includegraphics[width=0.95\linewidth]{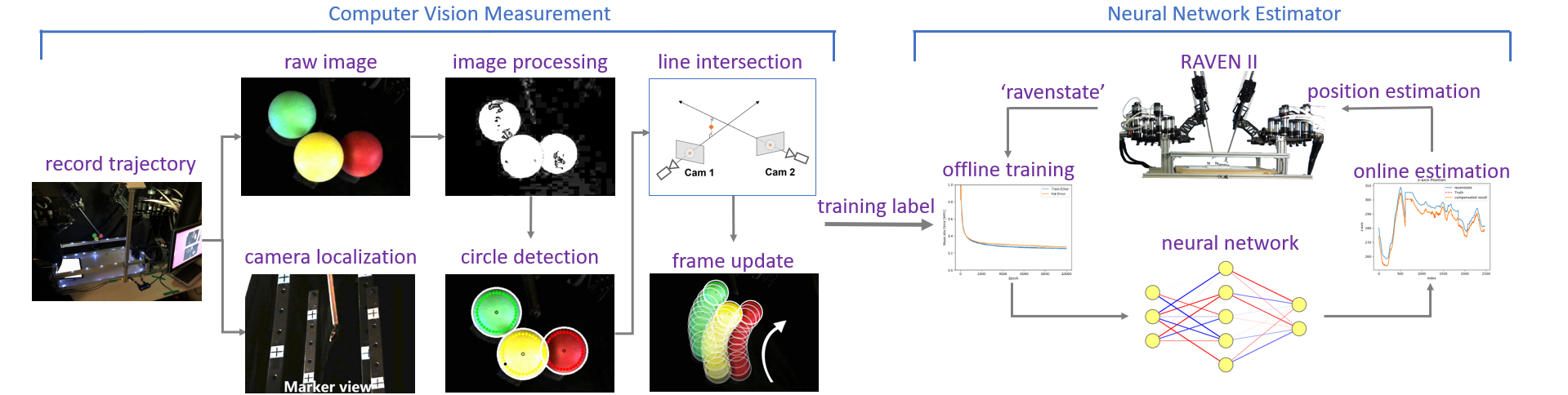}
  \caption{Workflow for proposed online position precision estimation system. Pipeline entails
     vision based ground truth measurement (left) and data driven robot position estimator (right). 
  Training labels passing from left to right are true RAVEN-II position information.}
  \label{fig:overview}
\end{figure*}

Haghighipanah et al. \cite{haghighipanah2015improving} proposed a joint level model-based approach for RAVEN-II pose estimation using an unscented Kalman filter \cite{wan2000unscented}. Although improved results were shown for the first three joints, the experiments were limited to repeatedly picking up a fixed mass. 
Also, joint pose estimates for the last four surgical tool joints were beyond the scope of the study and were suggested to be corrected through vision. In $2018$, Seita et al. \cite{seita2018fast} presented a data-driven Cartesian pose calibrator for the daVinci Research Kit experimental surgical robotic platform (dVRK) \cite{kazanzides2014open}. That study successfully achieved autonomous surgical debridement through a two-phase calibration procedure. The result shows high precision in both position and orientation, and is tailored to the specific surgical task of debridement. In $2014$, Mahler et al. \cite{mahler2014learning} used Gaussian process regression and data cleaning to reduce the error of RAVEN-II end effector, and by including the velocity information, the accuracy could be further improved.
\begin{figure}[ht!]
  \includegraphics[width=0.48\textwidth]{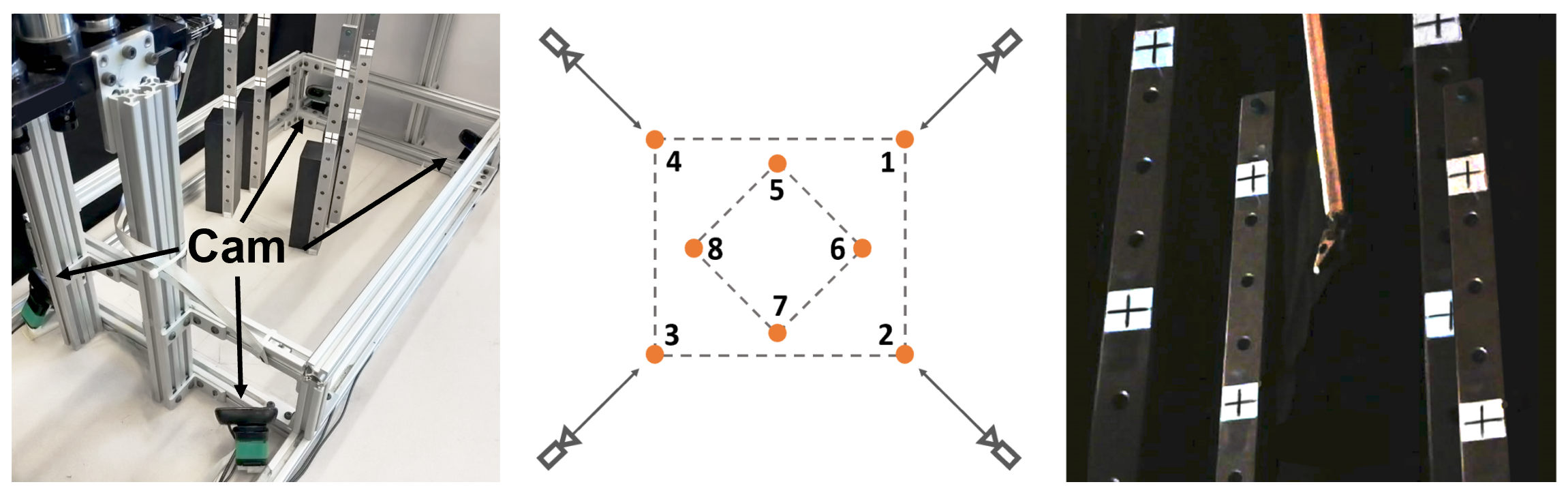}
  \caption{$16$ marker points used for camera pose calibration are placed at 8 ground locations in robot workspace.}
  \label{fig:marker-setup}
\end{figure}

\subsection{Contributions}
In this work, the authors built a robot position precision estimator on RAVEN-II that is not task-specific and spans the robot workspace. Thus, a pipeline was built to collect precise RAVEN-II position data traversing the workspace from teleoperation trials by human operators. The ground truth position was carefully derived through calibrated stereo vision. Finally, the dataset was used to train a neural network model that estimates position error. To the best of the authors' knowledge, this work is first to simultaneously
\begin{enumerate}
    \item introduce a cost-effective approach for vision-based precise robot position data collection;
    \item train a neural network model with 1000Hz detailed sensor and controller state information as input; 
    \item implement and analytically quantify the performance of a data-driven precision estimator of the robot position in the entire robot workspace.
\end{enumerate}

\section{Methods}
\subsection{System Workflow}
The online end effector pose estimation system consists of two phases - vision-based ground truth measurement and neural network estimator, as shown in Fig. \ref{fig:overview}.  
\subsubsection{Phase One}
$4$ webcams are mounted with poses determined by a two-phase calibration procedure (Fig. \ref{fig:marker-setup}, left). During data recording, $3$ distinct colored balls are fixed to the end effector as markers (Fig. \ref{fig:solve_end_effector_pose} left). The balls are only used to collect training data and are removed in real operation. In addition, the balls and the holder are hollow and very light, for a negligible load to the system. For each image frame, preprocessing steps are performed and followed by Hough circle detection \cite{yuen1990comparative}. A frame update algorithm is then adopted to prevent false positives by comparing circles detected in subsequent images. Finally, the $3$ circle centers yield ground truth end effector position up to $0.5$ mm accuracy.
\subsubsection{Phase Two}
There are $2$ stages in phase two - offline training and online estimation. The collected position data is used to train a neural network. Next, the trained neural network model provides online estimation of the end effector position with an average precision of around $0.8$ mm, a significant improvement from the RAVEN-II feedback position.

\begin{figure}[ht!]
  \includegraphics[width=0.45\textwidth]{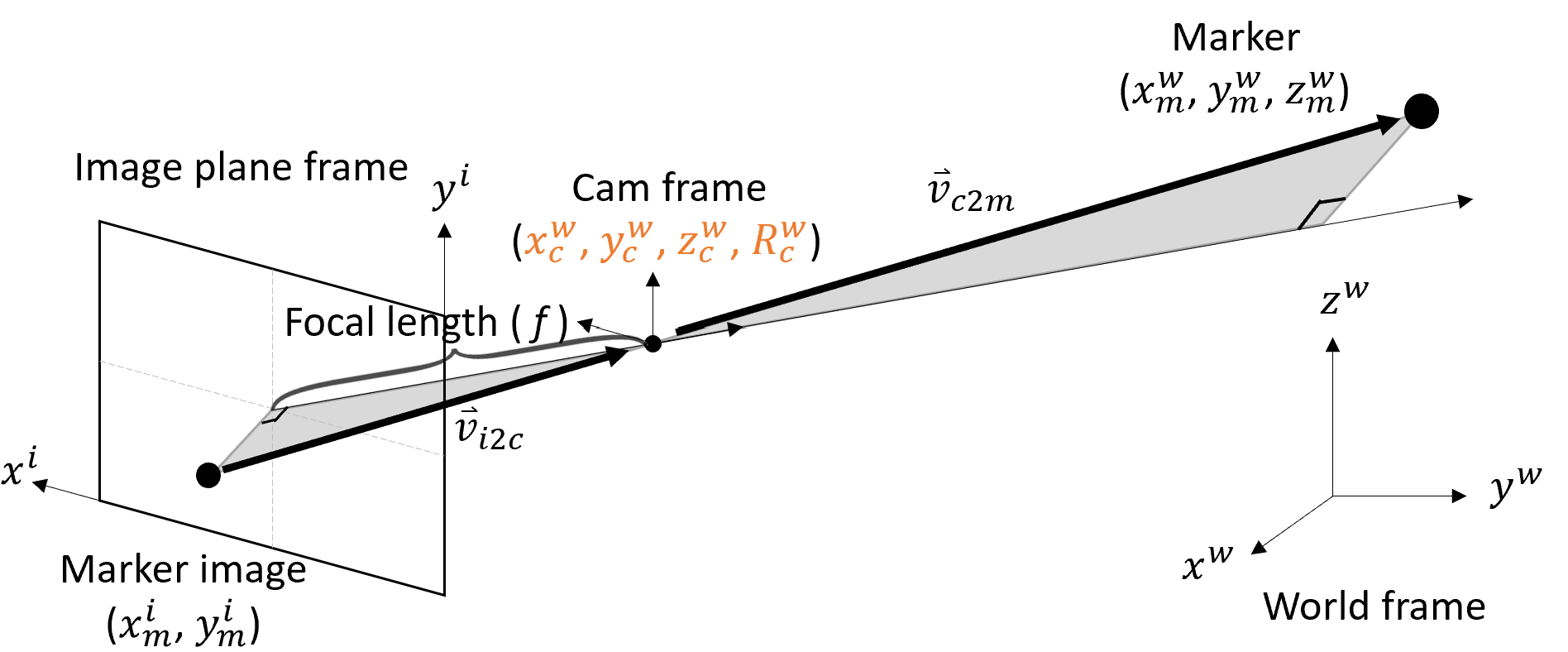}
  \caption{Geometric relations among image plane, camera and marker.}
  \label{fig:marker_equation}
\end{figure}
\subsection{Camera Localization}
\begin{figure*}[ht!]
  \centering
  \includegraphics[width=0.9\linewidth]{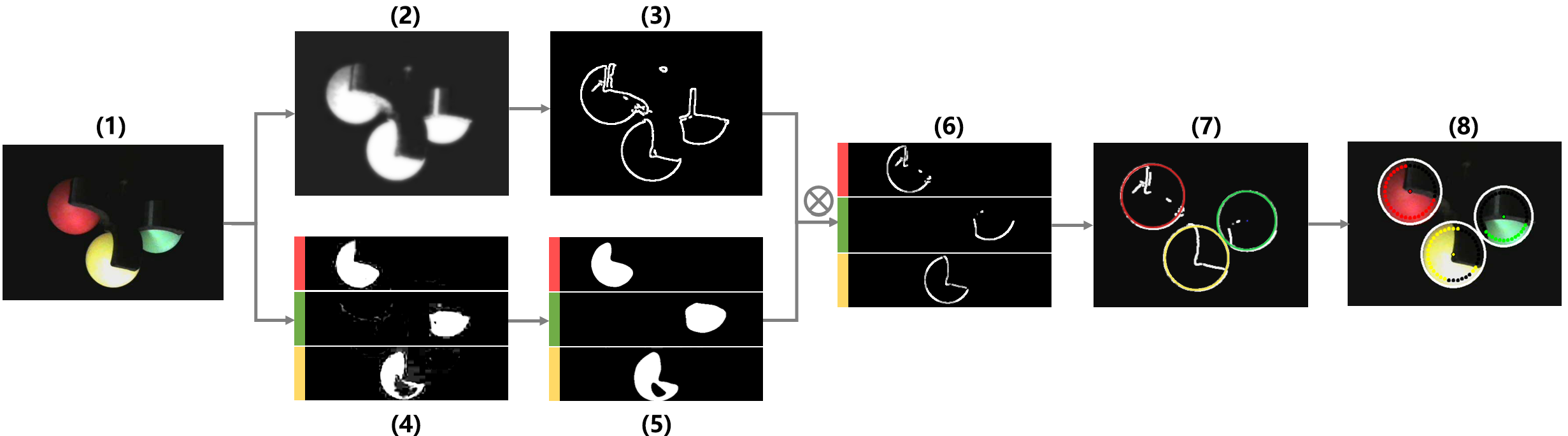}
  \caption{Circle detection method and the preprocessing steps. Color image (1) is first converted to gray and Gaussian blur is applied (2). Then Canny edge (3) is taken from the gray image. In the other pathway, the color image (1) is divided into red, green, and yellow color channels (4). Then Gaussian blur with  $\sigma$ of 10\% radius followed by binarization is applied (5). Image (6) is elementwise multiplication of (3) and (5). Hough circle detection is applied in each channel in image(6), which combines to result in (7). Finally, image (8) is a color check near the circle border to eliminate false positives.}
  \label{fig:img_process}
\end{figure*}
Since the goal of phase one is to obtain ground truth robot position through vision, accurate camera pose identification in relation to the robot frame is crucial. The existing camera extrinsic matrix solver in MATLAB \cite{fetic2012procedure} requires placing a chessboard at a known pose with respect to the world frame. The resultant camera pose from said solution contains error up to 40 millimeters if only 1-5 chessboards are used. Although placing more chessboards (usually more than 20) theoretically improves precision, the problem of acquiring accurate chessboard poses also becomes increasingly difficult due to the constrained workspace. 
As a practical alternative, 
we use $1$ chessboard and $16$ marker points together (Fig. \ref{fig:marker-setup}). 
The chessboard
provides a rough cameras pose estimate and the 16 markers further refine it. 


We define world frame $w$, chessboard frame $b$, camera frame $c$ and image frame $i$.
The camera position and orientation with respect to the world coordinate are respectively 
${p_{c}^{w}}=[{x_{c}^{w}} \: {y_{c}^{w}} \: 
{z_{c}^{w}}]^T$ and ${R_{c}^{w}}\in\mathbb{R}^{(3\times3)}$.

Similarly, the marker position in the world coordinate is \({p_{m}^{w}}=[x_m^w \: y_m^w \: z_m^w]^T\) whereas its 2D projection on the image frame is \({p_{m}^{i}}=[x_m^i \: y_m^i]^T\). Let distance vectors
\begin{eqnarray}
    \vec{v}_{c2m}&=&{p_{m}^{w}}-{p_{c}^{w}}\,,\\
	\vec{v}_{i2c}&=&R_{c}^w
	\begin{bmatrix}
	p_{m}^{i} \\
	f
	\end{bmatrix}
	=R_{b}^w \cdot R_{c}^{b}
	\begin{bmatrix}
	x_m^i \\
	y_m^i \\
	f
	\end{bmatrix}
	\,,
	\label{eqn:vector_i_cam}
\end{eqnarray}
where $f$ is the camera focal length. According to the geometric relations illustrated in Fig. \ref{fig:marker_equation}, the following three equations hold true for each marker from each camera view:

\begin{eqnarray}
	\frac{\vec{v}_{c2m}(k)}{\vec{v}_{i2c}(k)}=\frac{\left\Vert \vec{v}_{c2m} \right\Vert_2}{\left\Vert \vec{v}_{i2c} \right\Vert_2}\quad \forall k\in[1,2,3] \,,
	\label{key-eq}
\end{eqnarray}
where $\vec{v}(k)$ represents the $k$th entry in the distance vector.

There are a total of $6$ unknown variables in ${p_{c}^{w}}$ and $R_{c}^{w}$. With $16$ equation sets (\ref{key-eq}) yield from all $16$ marker points, the Levenberg-Marquardt algorithm\cite{more1978levenberg} finds optimal camera pose with initial guess \({p_{c}^{w}}'\) and \({R_{c}^{w}}'\) from the chessboard.

\subsection{Auto Hough Circle Detection}
Localization of the colored balls assists with end effector position acquisition.  2D circle detection is the first step to localize colored balls. Hough circle detection \cite{yuen1990comparative} is chosen due to its robustness to occlusion. Yet, it is sensitive to bright edge noises, so heuristic parameter tuning and image enhancements are necessary. Fig. \ref{fig:img_process} shows the pre-processing steps with  the following  design details:
\paragraph{Edge Detection (1)-(3)} These steps increase the signal to noise ratio and clear up noisy edges.
\paragraph{Color Segmentation (1)-(4)-(5)} After this, circle segments in the color channels are intentionally enlarged.
\paragraph{Border Refinement (3)(5)-(6)}Since circles in b) are larger than a). The circle canny edges \cite{canny1987computational} will be kept, and any edges outside the circles will be removed.
\paragraph{Circle Identification (6)-(7)-(8)} After circles are detected, sample points near the border are used to ensure the circle color matches the target color with similarity more than $25\%$. Otherwise, the circle is considered a false detection.

\begin{figure}[ht!]
  \includegraphics[width=0.48\textwidth]{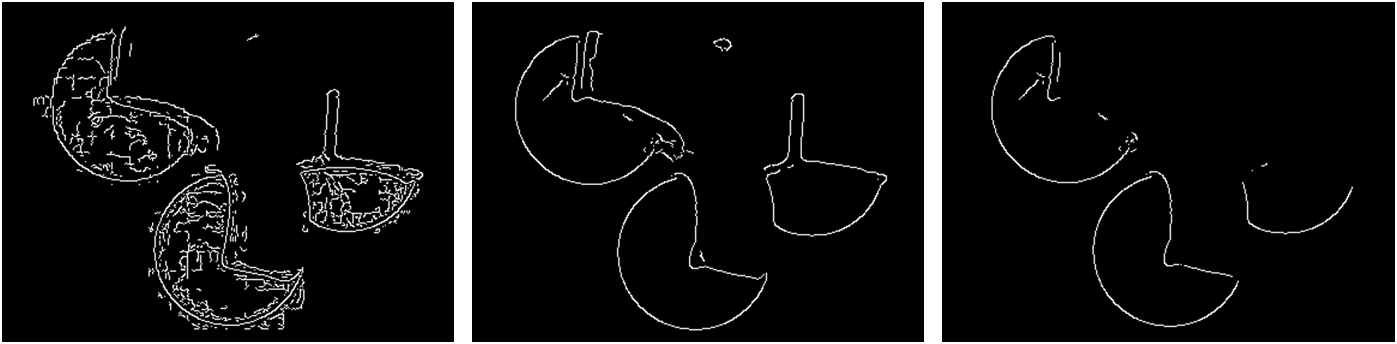}
  \caption{Comparison of Canny edge on original image (left), normal blurred image (middle) and processed image by our system (right). }
  \label{fig:canny_comparation}
\end{figure}

The `HoughCircle' function in OpenCV \cite{bradski2008learning} is used to detect circles.
Two heuristically chosen parameters are inverse ratio $dp=1$ and high Canny
threshold $para1=100$. To improve adaptability, an automated parameter tuning algorithm is
designed for the accumulator threshold $para2$, where small values indicate a higher chance
of false positives. 

Suppose $k$ circles are expected, the algorithm uses bisection to approach the lower bound of $para2$, such that `HoughCircle' comes close to returning $(k+1)$ circles but still returns $k$. In the case where $para2$ jumps between $(k-1)$ and $(k+1)$, increasing $\sigma$ in Gaussian blur helps, but is conducted only if necessary - due to decreased precision.

\subsection{Frame Update Algorithm}

There are $2$ main purposes of the frame update algorithm. The first is to decide if any camera is returning false circles. 
The second is to provide other parameters to `HoughCircle', including the minimum distance between circle centers $d_{min}$, and extrema of circle radius $r_{min}$ and $r_{max}$. 
Setting a low $d_{min}$ and wide range between $r_{min}$ and $r_{max}$ allows more chance to detect circles, but it also increases computational cost and the risk of false positives. Thus, under smooth end effector motion, $d_{min}$, $r_{min}$ and $r_{max}$ values are bounded by the detection result from the previous frames.

There are a total of $4$ cameras. For each ball, there should ideally be $4$ circles detected - one corresponding to each of the $4$ camera views. After the frame update algorithm, each detected circle is proclaimed as either effective or suspended. 

Every circle is initialized as effective. If any of the conditions below holds true, an effective circle will be suspended:

\begin{itemize}
\item The movement of the circle center in successive frames exceeds a predetermined motion threshold.
\item Color check in fig. \ref{fig:img_process} (8) fails to match target color by $25\%$.
\item Hough circle detection does not return a circle, even with the largest tolerable $\sigma$ for Gaussian blur.
\end{itemize}

On the other hand, if all the following conditions are satisfied for more than 5 frames, a suspended circle will become effective:

\begin{itemize}
\item The Hough circle returns result normally.
\item Color check in fig. \ref{fig:img_process} (8) is successful.
\item The 3D reconstructed ball center position from effectively detected 2D circles is consistent with that of this suspended 2D circle information.
\end{itemize}

A line that connects a 2D effective circle center and its associated camera center forms a 3D ray. The 3D position of the ball center is the intersection point of rays. 
These rays usually do not intersect perfectly, and the midpoint of the common perpendicular of the two rays is selected to be the intersection point.
When the number of effective circles is larger than $2$, the mean position is taken from each pair of rays. Thus, $2$ or more effective circles of the same ball must be detected at any time instance for successful localization. If not, the system skips a few frames and restarts, all while raising the tolerable range for the parameters. Once there are enough effective circles, the parameter tolerance converges back down.

\subsection{End Effector Localization}

\begin{figure}[ht!]
  \includegraphics[width=0.48\textwidth]{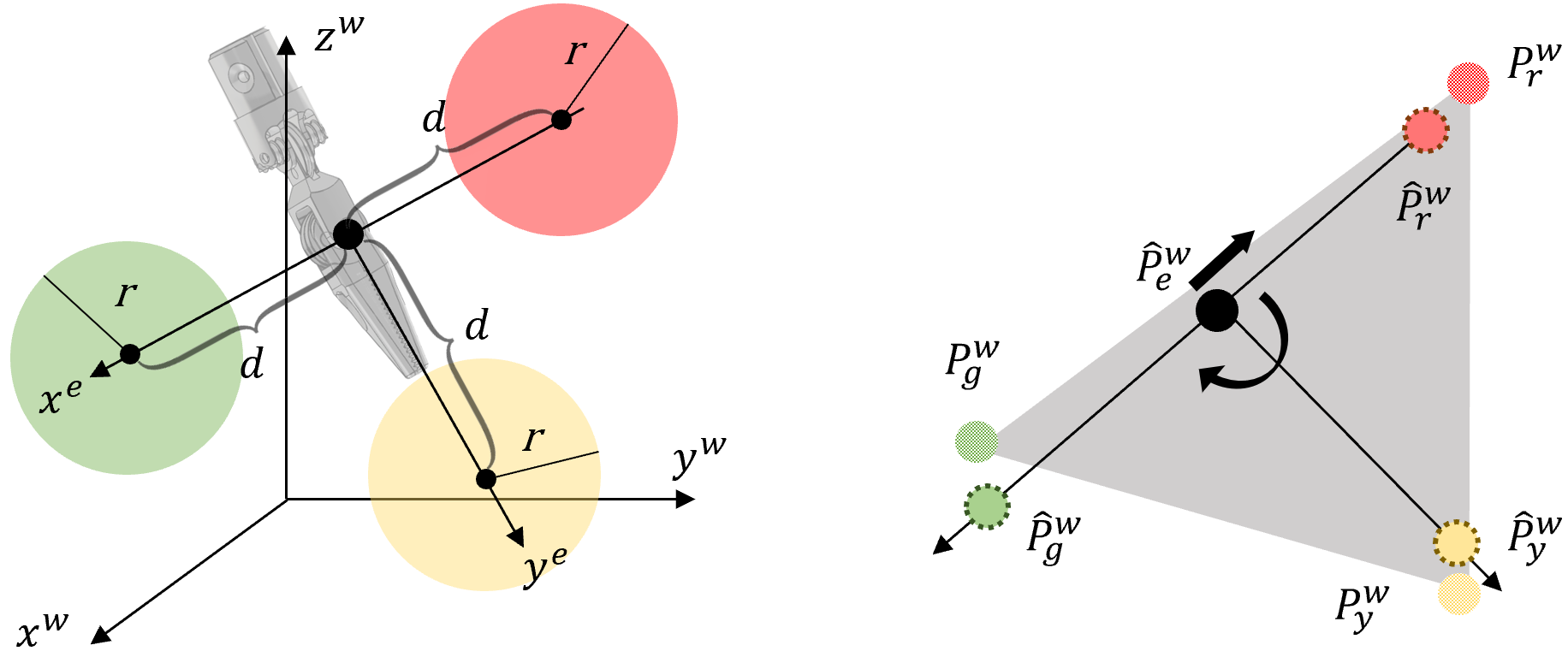}
  \caption{Geometric relations between ball centers and end effector position.}
  \label{fig:solve_end_effector_pose}
\end{figure}

As Fig. \ref{fig:solve_end_effector_pose} shows, the $3$ colored balls are fixed around the end effector center (the black point, which is defined by the RAVEN-II system). The end effector center position shares the same plane with the $3$ ball centers.  The distances between the end effector center and each ball center are defined as $d=38$ mm and each ball has radius $r=20mm$. An end effector coordinate frame is designed for RAVEN-II, where the origin is the end effector point, the X axis points toward the green ball center, and the Y axis points at the yellow ball center, which is also  co-linear with RAVEN-II axis $X_{6b}$\cite{king2012kinematic}.

The detected ball centers in the world coordinate frame are: green $p_{g}^w=[x_{g}^w \: y_{g}^w \: z_{g}^w]^T$, yellow $p_{y}^w=[x_{y}^w \: y_{y}^w \: z_{y}^w]^T$ and red $p_{r}^w=[x_{r}^w \: y_{r}^w \: z_{r}^w]^T$. These $3$ points are used to estimate the end effector center $ \hat{p}_e^w=[\hat{x}_e^w \: \hat{y}_e^w \: \hat{z}_e^w]^T$ and the orientation $\hat{R}_e^w = Rot(z,\hat{\gamma})\cdot Rot(y,\hat{\beta})\cdot Rot(x,\hat{\alpha})$ represented in Euler angles. From the setup of the end effector and three colored balls, the estimated ball centers can be represented by
\begin{eqnarray}
	\hat{p}_{g}^w&=&\hat{R}_e^w \cdot 
	\begin{bmatrix}
	d & 0 & 0
	\end{bmatrix}
	^T + \hat{p}_e^w \\
	\hat{p}_{y}^w&=&\hat{R}_e^w \cdot 
	\begin{bmatrix}
	0 & d & 0
	\end{bmatrix}
	^T + \hat{p}_e^w \\
	\hat{p}_{r}^w&=&\hat{R}_e^w \cdot 
	\begin{bmatrix}
	-d & 0 & 0
	\end{bmatrix}
	^T + \hat{p}_e^w
\end{eqnarray}
To solve for the end effector pose, a cost function $C$ with $6$ unknown variables $(\hat{x}_e^w, \hat{y}_e^w, \hat{z}_e^w,\hat{\alpha},\hat{\beta},\hat{\gamma})$ can be set up. Minimizing the cost gives an optimal solution of the unknown variables, 
\begin{eqnarray}
    C=(p_g^w-\hat{p}_g^w)^2 + (p_y^w-\hat{p}_y^w)^2 + (p_r^w-\hat{p}_r^w)^2
\end{eqnarray}
Take the middle point of $p_g^w$ and $p_r^w$ as the initial guess of $\hat{p}_e^w=[\hat{x}_e^w \: \hat{y}_e^w \: \hat{z}_e^w]^T$, denoted as $P_{e0}^w$. The initial guess of orientation is taken by setting $\overrightarrow{P_{e0}^w P_g^w}$, $\overrightarrow{P_{e0}^w P_y^w}$ as x and y axis of the end effectors frame. Then, the Nelder-Mead Simplex Method\cite{lagarias1998convergence} is used to find the optimal value of $(\hat{x}_e^w, \hat{y}_e^w, \hat{z}_e^w,\hat{\alpha},\hat{\beta},\hat{\gamma})$, which is the pose of the end effector.


\subsection{Neural Network Architecture}
The training data for the neural network is represented as:
\begin{eqnarray}
	\chi_{NN} = \left\{\left(\text{ravenstate}^{(k)},\text{err}^{(k)}\right)\:|\:k=[1...N]\right\} \,,
\end{eqnarray}
where `ravenstate' is a ROS topic that contains real-time kinematics and dynamic information of RAVEN-II. And `err' is the difference between the RAVEN-II reported end effector positions and the ground truth collected through vision-based measurement in phase one. More details follow:
\subsubsection{The Features}
The information provided in a `ravenstate' message includes: (a) kinematics derived Cartesian pose, (b) desired Cartesian pose, (c) current joint pose, (d) desired joint pose, (e) motor and joint velocities, (f) motor torques, (g) desired grasper pose and other information, all of which are added as features of the neural network.

\subsubsection{The Labels}
We chose the end effector {\it position error} as the label instead of the ground truth position itself because RAVEN-II already has end effector pose feedback in the `ravenstate' topic. To get a more accurate estimation, one only needs to correct RAVEN-II pose feedback derived from joint pose and forward kinematics, instead of creating a new estimation. In fact, this design is more robust to static Cartesian positional offsets contained in the `ravenstate' reported position, which slightly differs every time the system restarts.
\begin{figure}[ht!]
  \includegraphics[width=0.48\textwidth]{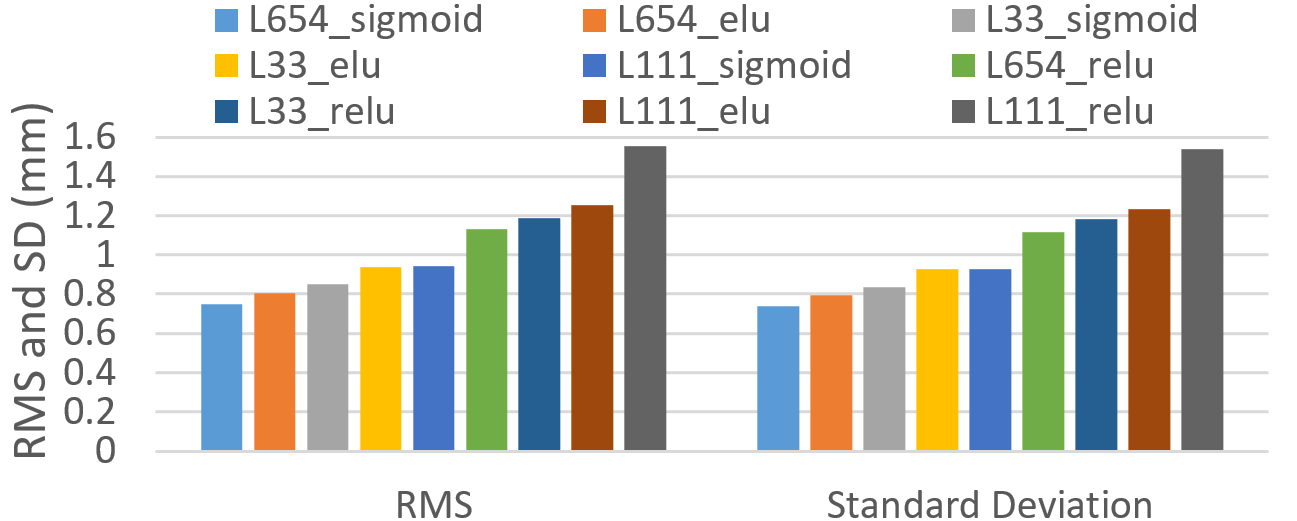}
  \caption{The performance of different hyperparameter sets. The legend `L654\_sigmoid' means that there are 3 hidden layers of 600, 500 and 400 units with sigmoid activation. `\_elu/\_relu' refers to ELU\cite{clevert2015fast} and RELU\cite{nair2010rectified} activation functions}
  \label{fig:network-param}
\end{figure}

\subsubsection{Network Structure and Parameters}
In order to determine the structure and hyperparameters of the neural network, 
the network was first trained with randomly chosen hyperparameters from a large range and then the hyperparameters narrowed to a smaller range. 
We evaluated more than 100 sets of hyperparameter values. Nine illustrative values are plotted in Fig. \ref{fig:network-param}.  
The following hyperparameter values yielded the best obtained performance:

\begin{itemize}
\item $3$ dense sigmoid layers of $600$, $500$, $400$ units.
\item batch normalization\cite{ioffe2015batch}, batch size = $1024$.
\item learning rate = $1\times10^{-8}$, epochs = $10000$.
\item regularization rate = $5\times10^{-6}(L_1)$.
\item Adam \cite{kingma2014adam} optimizer: ($\beta_1=0.9, \beta_2=0.999, \epsilon=10^{-8}$)
\end{itemize} 


\section{Experimental Result}
\subsection{Experiment Setup}
A standard RAVEN II surgical robot was used with a remote controller. 
Only the left arm was activated and there was no control signal sent to the right arm. 
The vision-based ground truth measurement system was set up and the environment was surrounded by black cloth to reduce background interference in images. 
To get data for neural network training, RAVEN-II was manually operated by a remote 
controller and moved randomly in the workspace for $140$ minutes. 
The data was recorded as time-synchronized pairs of `ravenstate' and ground truth end 
effector positions. The recorded trajectories contained a total of $49,407$ data pairs, in which each ravenstate consists of 118 floats for each arm.
\[
\left( \text{ravenstate}^{(k)},\text{err}^{(k)}\right)   k=1...49,407 \,.
\]
\begin{figure}[ht!]
\centering
  \includegraphics[width=0.45\textwidth, trim={0mm 10mm 0mm 0mm},clip]{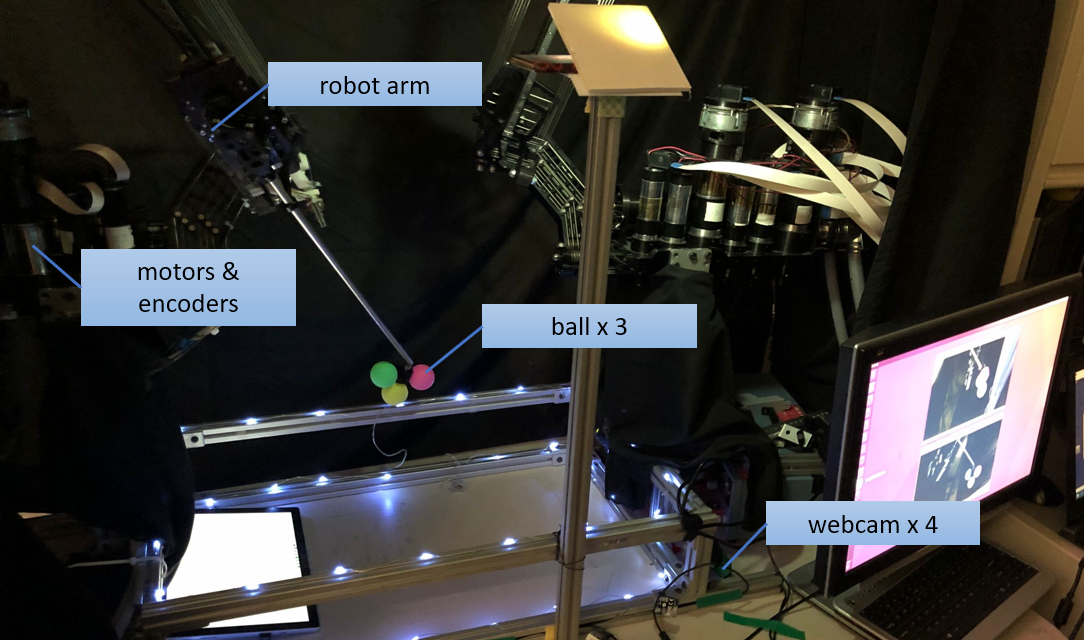}
  \caption{Experimental setup.}
  \label{fig:experiment-setup}
\end{figure}

\subsection{Neural Network Offline Training}
The training data $\chi_{NN}$ was used to train the neural network $f_{NN}: \mathbb{R}^{118} \rightarrow  \mathbb{R}^3$ to map input `ravenstate'$^{(k)}$ to the output end effector position error $err^{(k)}$. Suppose $p_e^w$ and $\hat{p}_e^w$ are the true and `ravenstate' reported RAVEN-II positions. After training, an online end effector position precision estimator was built, where $p_e^w=\hat{p}_e^w+err$.

The training data comes from recorded trajectories, which are a series of continuous points traversing most of the workspace.
However, the neural network estimator is expected to work in the entire workspace. The technique of randomly choosing subsets of points in the recorded trajectory to 
form the validation set and test sets might only achieve high performance on the training set trajectories, and rapidly decreases its accuracy when the robot leaves the trained trajectories. After several tests, it was found that to prevent poor performance during online estimation at unseen points due to overfitting, the validation set and test set should be carefully selected. In particular, the validation set and test set are randomly chosen {\it trajectories} in the dataset, instead of random sample points. The purpose is to make sure that the trained neural network has similar accuracy in the whole workspace instead of performing well only along the recorded trajectories. 

\subsection{Performance Analysis}

First, the accuracy of the measurement system was tested. The error of measurement mainly comes from two aspects: 1) inaccurate circle detection in the images; 2) inaccurate camera localization.
The $1^{st}$ type of error was calculated by assuming camera localization is perfect. $200$ frames were chosen, and the centers of the circles were manually marked in the images. 
Then, the ball centers were calculated using manually marked circle centers and the result 
was used as the ground truth for accuracy analysis. 
The $2^{nd}$ type of error was calculated by assuming circle detection was perfect. 
$16$ marker points with known position inside the workspace were chosen and manually marked 
in the image plane. 
After ball centers were detected, the optimization method introduced in II.E was used to solve the end effector position and further improved the accuracy.
Finally, the result was compared with the ground truth and analytically shown in TABLE \ref{table:CV_error}.

\begin{table}[th!]
  \caption{The Error of Measurement System}
  \label{table:CV_error}
  \centering
  \begin{tabular}{lllll}
    \toprule
   Axis  & \multicolumn{2}{l}{Single Ball Center} & \multicolumn{2}{l}{End Effector Position}\\
    \midrule
     	&RMS(mm)&SD(mm)&RMS(mm)&SD(mm)\\
    \midrule
    x   & 0.6273&0.2891 & 0.4229 & 0.0513\\
    y   & 0.6354&0.3000 & 0.2896 & 0.0949\\
    z   & 0.3004&0.1029 & 0.1521 & 0.0718\\
    \midrule
    3D        &0.9866&0.2315 & 0.5346 & 0.0695\\
    \bottomrule
  \end{tabular}
\end{table}

After training the neural network offline, an online neural network estimator was built, 
which took the `ravenstate' as input and output the estimated end effector position, 
$e^w=\hat{p}_e^w+err$, 
which is roughly $10$ times more accurate than the RAVEN-II original estimation calculated by kinematics. The test set, consisting of $10\%$ of the recorded data, was untouched during all training procedures, and was used to test the system performance.

\begin{figure*}[ht!]
  \centering
  \includegraphics[width=0.95\linewidth]{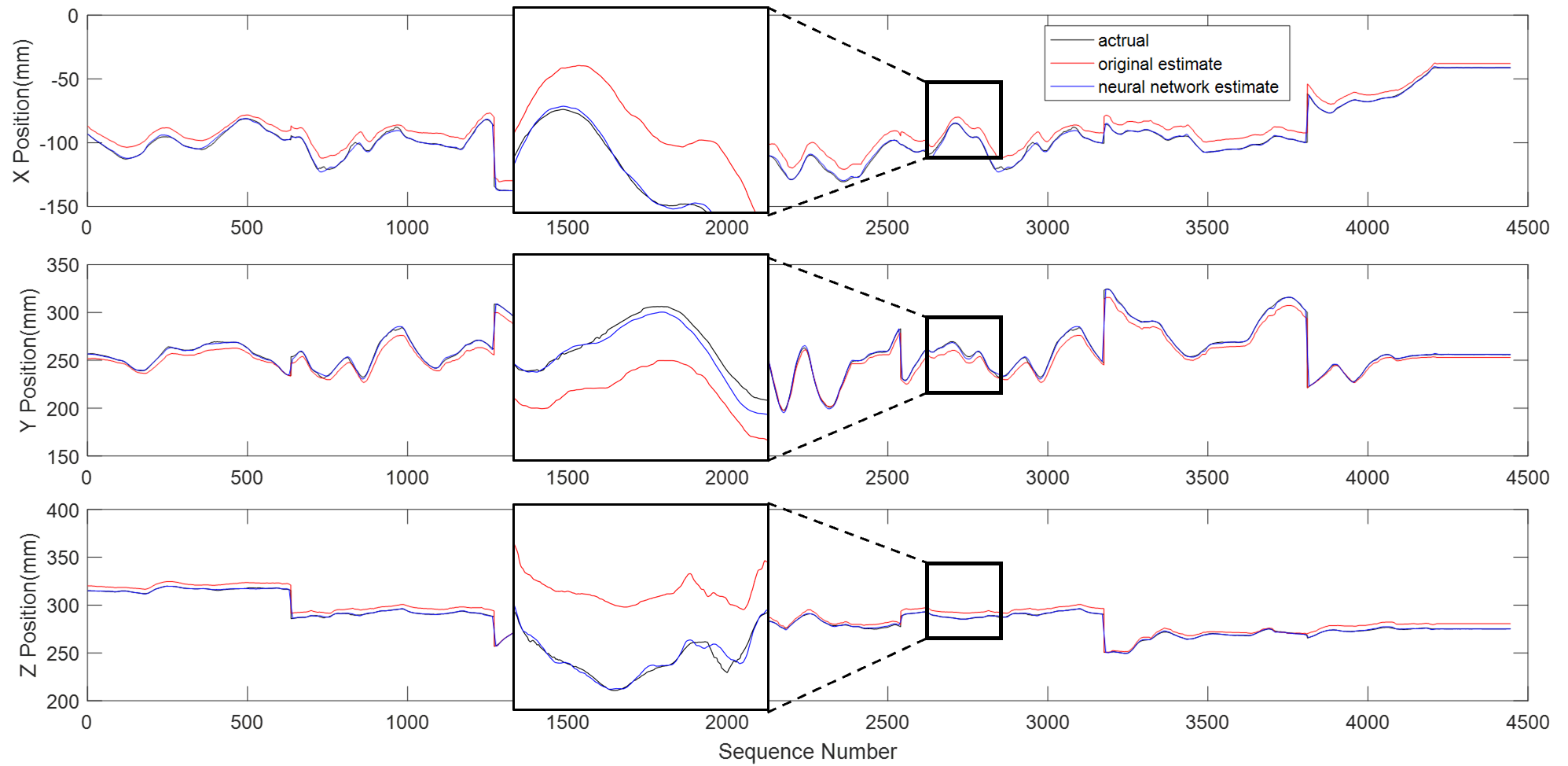}
  \caption{Comparison of actual position, motor encoder based forward kinematics position, and neural network estimation. The neural network estimator decreases the positional RMS error by 83.6\% and the standard deviation of error by 59.4\%.}
  \label{fig:original_kinematics_estimation_and_neural_network}
\end{figure*}

\begin{figure}[ht!]
  \includegraphics[width=0.48\textwidth,]{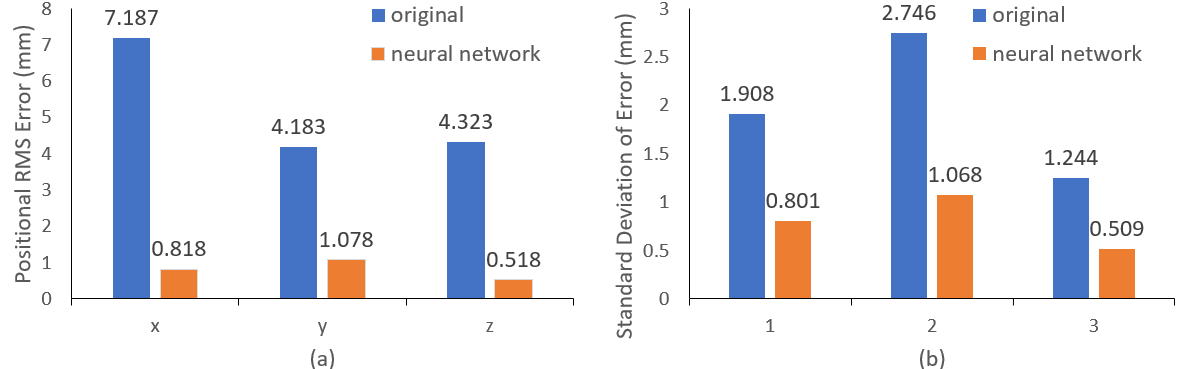}
  \caption{RMS error and standard deviation of error in end effector position with and without neural network online estimator.}
  \label{fig:RMS_error_and_STD_error}
\end{figure}

Fig. \ref{fig:original_kinematics_estimation_and_neural_network} shows the online estimation result. 
The actual value was obtained using the measurement system, and the original estimate came
from the RAVEN-II forward kinematics, containing errors from cable-driven mechanism \textit{etc}. Fig. \ref{fig:RMS_error_and_STD_error} shows the RMS error and standard deviation of the error. 

\section{Discussions and Future Work}
\subsection{Performance of Phase One}
The vision-based ground truth position measurement system was used to measure the RAVEN-II end effector position, 
but can be extended to measuring other object poses as long as $3$ colored balls can be fixed on the object. 
Since there are $4$ cameras around the workspace viewing the large radius balls, the system is robust 
to occlusion. The measurement system achieves accuracy below $0.5$ mm in each axis. (TABLE.\ref{table:CV_error})

In summary, the proposed procedure for
acquiring ground truth position is a cost-effective option to satisfy the required accuracy and can further improve 
precision by increasing the ball count or using higher resolution cameras.

As illustrated in Fig. \ref{fig:original_kinematics_estimation_and_neural_network}, the data was collected through manual teleoperation of RAVEN-II for $140$ minutes through random trajectories traversing the robot workspace. 
Compared to preprogrammed periodic trajectories, this dataset provides data 
spanning most of the
workspace and the motion pattern is more realistic to teleoperated surgical operations. 

\subsection{Advantages of Phase Two}
After offline training, we built an online estimator, which gave a significantly more accurate end effector position estimate. Since the neural network input `ravenstate' is an existing ROS topic from 
RAVEN-II, the online estimator requires no additional sensor or information and can be applied to any 
RAVEN-II robot. However, with continued use, the model may need to be retrained due to changes in
cable tension and wear for individual robots. 

Fig. \ref{fig:RMS_error_and_STD_error} shows that the neural network estimator can not only decrease the RMS error, 
but also the standard deviation of the error, which means that the online estimator can improve precision 
beyond applying a static Cartesian offset alone. Moreover, our estimator could potentially be utilized 
in other robots where transmission compliance and losses are significant influences on precision. 
In the future, tests will be conducted to identify which information in the `ravenstate' ROS topic is more 
impactful in the estimator. This could help further improve the model by trimming down features with less 
influence on end effector position prediction.

\section{Conclusion}
Due to compliance and losses in the transmission mechanism (cable/pulley links in the RAVEN-II), 
indirect measurement of joints and other external uncertainties, estimation of 
the precise end effector position is challenging. In this work, a cost-effective online RAVEN-II position precision 
estimator is implemented and tested on a 140-minute trajectory set. 
The system  entails a vision-based ground truth position measurement system and an online data-driven position 
estimator based on a neural network. Although the total cost of the measurement is around one hundred dollars (mostly the cost of four webcams),
the sub-millimeter accuracy achieved is more than $10$ times better than the RAVEN-II  position accuracy based on 
motor-mounted encoders. 
The neural network estimator decreases the positional RMS error by $83.6\%$ and the standard deviation of error by $59.4\%$. 
Furthermore, the estimator requires no additional sensors or information other than the RAVEN-II built-in `ravenstate' ROS topic (updated at 1000Hz), which contains kinematic and dynamic information of RAVEN-II. 
Finally, the proposed cost-effective position estimator can be generalized to other RAVEN-II
sites, as well as other robots with accuracy affected by compliance and losses in transmission elements between motors and joints. 
Although robotic surgeons today readily compensate for imperfect position control, as commercial  surgical robots incorporate human augmentation and autonomous functions, the need for accurate position estimate and control will increase. 

\addtolength{\textheight}{-1cm}
\bibliographystyle{IEEEtran}
\bibliography{ref}

\end{document}